\documentclass[11pt]{article}

\usepackage[a4paper,margin=1in]{geometry}
\usepackage{amsmath,amssymb,amsthm}
\usepackage{graphicx}
\usepackage{booktabs}
\usepackage{algorithm}
\usepackage{algpseudocode}
\usepackage{natbib}
\usepackage{hyperref}
\usepackage{xcolor}
\usepackage{enumitem}
\usepackage{caption}
\usepackage{subcaption}
\usepackage{bm}
\usepackage{mathtools}
\usepackage{float}

\hypersetup{
    colorlinks=true,
    linkcolor=blue!50!black,
    citecolor=blue!50!black,
    urlcolor=blue!50!black
}

\newtheorem{theorem}{Theorem}[section]

\newtheorem{definition}[theorem]{Definition}
\newtheorem{remark}[theorem]{Remark}

\newcommand{\R}{\mathbb{R}}
\newcommand{\C}{\mathbb{C}}
\newcommand{\CP}{\mathbb{CP}}
\newcommand{\M}{\mathcal{M}}
\newcommand{\T}{\mathcal{T}}
\newcommand{\Exp}{\mathrm{Exp}}
\newcommand{\Log}{\mathrm{Log}}
\newcommand{\dist}{d}
\newcommand{\inner}[2]{\langle #1, #2 \rangle}
\newcommand{\norm}[1]{\lVert #1 \rVert}
\newcommand{\preshape}{\mathcal{S}}
\DeclareMathOperator*{\argmin}{arg\,min}

\title{\textbf{Riemannian Shape Analysis of the Corpus Callosum in Kendall Space: Aging and Alzheimer's Disease}}

\author{
\begin{minipage}{0.45\textwidth}
\centering
Olakunle S. Abawonse$^{1}$\\
\small African Institute for Mathematical Sciences (AIMS), Rwanda\\
\small \texttt{olakunle.abawonse@aims.ac.rw}
\end{minipage}
\hfill
\begin{minipage}{0.45\textwidth}
\centering
Fatou Fall$^{2}$\\
\small African Institute for Mathematical Sciences (AIMS), Senegal\\
\small \texttt{fatou.fall@aims-senegal.org}
\end{minipage}
}

\date{}

\begin{document}

\maketitle

\begin{abstract}
The corpus callosum (CC) is a major white-matter structure and a well-established marker of brain aging, but most studies quantify it using scalar summaries that discard its boundary geometry. We present a Riemannian shape-space framework \cite{fletcher2013geodesic-regression} for analyzing age-related morphological change in the midsagittal CC, applied to the OASIS-1 cohort. Each contour is represented by $128$ landmarks and embedded into Kendall shape space, where translation, rotation, and scale are removed. We derive a multivariate geodesic regression with exact Riemannian gradients and use the fitted age-velocity field to localize age-related deformation to five anatomical sub-regions. In the cognitively normal cohort ($n = 252$), geodesic regression outperforms the Euclidean linear benchmark ($R^2 = 0.1355$ vs.\ $0.1216$). Regional energy is anterior-dominant: the Genu carries $46.1\%$ and the Anterior Body $23.3\%$ of total age-related shape change, together accounting for $\sim 69\%$ despite comprising only $\sim 41\%$ of landmarks. Signed projections confirm the ordering (Genu $r = 0.562$; Anterior Body $r = 0.507$). In contrast, age explains less than $0.5\%$ of shape variance in Alzheimer's disease ($n = 88$), indicating that the disease disrupts the healthy aging trajectory. A tangent-space classifier achieves an age-group AUC of $0.791$ from the 2D contour alone, exceeding a recent volumetric benchmark ($0.67$).

\vspace{0.3cm}
\end{abstract}

\section{Introduction and Problem Statement}
%==================================================================

\subsection{Motivation}

The corpus callosum (CC) is the largest white-matter fiber bundle in the human brain, connecting homologous regions of the left and right hemispheres. Its compact midsagittal profile makes it a standard target for studying neurodevelopment, aging, and neurodegenerative disease \citep{hampel1998corpus, luders2010bell}. Most existing studies quantify CC changes using scalar summaries—total area, sub-regional partitions, or volume normalized by intracranial volume. Such measurements are simple to compute but discard the two-dimensional geometric configuration of the boundary: two corpora callosa with identical area can differ substantially in curvature, elongation, and local bending.

\emph{Shape}, in Kendall's sense \citep{kendall1984shape}, is the geometric information that remains after translation, rotation, and scaling have been removed. This induces a curved, non-Euclidean space with a well-defined Riemannian metric, in which the ``pure shape'' of an anatomical contour can be represented, compared, averaged, and statistically modeled without contamination from pose or size.

The distinction between size and shape matters: size and shape can evolve
along different lifespan trajectories, and the spatial pattern of
age-related change across the corpus callosum is itself a subject of
active investigation. A substantial body of work has established an
\textbf{anterior-to-posterior gradient} of age-related vulnerability, in
which the genu and anterior body—connecting the frontal lobes—show the
most pronounced microstructural and morphometric change. Payas et
al.~\cite{payas2026region} used tractography and machine learning to
demonstrate that \emph{``the CC does not age uniformly; instead, it shows
a region-specific pattern in which anterior and central pathways are more
susceptible to microstructural change, whereas posterior pathways remain
comparatively stable in adulthood.''} Liu et
al.~\cite{liu2021connectivity}, in a study of 1{,}086 healthy adults,
similarly reported an \emph{``overall `anterior-to-posterior'
region-specific changing trend during aging,''} with anterior CC
subregions exhibiting larger microstructural and connectivity changes.
Sullivan et al.~\cite{sullivan2010quantitative} framed this as a
consensus in the field, stating that \emph{``a current consensus is that
in adults advancing age affects anterior brain regions disproportionately
more than posterior regions,''} and reported paired comparisons in which
\emph{``FA was lower and diffusivity higher in anterior than posterior
fibers''} of the genu versus the splenium. Fan et
al.~\cite{fan2019age}, using high-gradient diffusion MRI, found that
age-related axonal alterations were \emph{``most pronounced in the
genu''} and \emph{``relatively absent in the splenium.''} Whether a
size-normalized shape representation can recover this anterior-dominant
structure cleanly is therefore both methodologically and biologically
meaningful.

\subsection{Problem Statement}

The central question addressed in this paper is:

\begin{quote}
\emph{Does the intrinsic Riemannian shape of the midsagittal corpus callosum contour, extracted from MRI, carry statistically exploitable information for characterizing healthy aging and for distinguishing age groups, beyond what is captured by scalar volumetric measures --- and can this framework localize the age-related geometric change to specific anatomical sub-regions?}
\end{quote}

To answer this question we pursue four specific, interrelated objectives:

\begin{enumerate}[label=(O\arabic*)]
    \item \textbf{Representation}: Represent the CC contour of each subject as a configuration of $k=128$ ordered planar landmarks, and embed the resulting shapes into Kendall's shape space $\Sigma_2^k \cong \CP^{k-2}$, a complex projective space of (real) dimension $2k-4 = 252$, i.e.\ complex dimension $126$.
    \item \textbf{Statistical modeling on the manifold}: Use the Riemannian logarithm map at the Fr\'echet (Karcher) mean shape to linearize the space (tangent-space PCA), and use geodesic regression to model shape as a function of age, contrasting it with ordinary (Euclidean) linear regression on landmark coordinates. Extend the single-covariate model to a fully multivariate geodesic regression that accommodates age, sex, and diagnosis jointly, mirroring multiple linear regression.
    \item \textbf{Regional decomposition}: Partition the 128 landmarks into five Witelson-style anatomical sub-regions using arc-length fractional boundaries anchored at the genu and splenium tips, and quantify how much of the fitted age-velocity field is concentrated in each region through a size-normalized kinetic energy decomposition.
    \item \textbf{Classification}: Using the tangent-space PCA coordinates as features, train and evaluate several classifiers (Logistic Regression, Random Forest, SVM with RBF kernel) to discriminate age groups, and benchmark the resulting discriminative power (AUC) against the volumetric approach of \citep{soysal2025machine}.
\end{enumerate}

\subsection{Contributions}

\begin{itemize}
    \item We present an end-to-end Riemannian pipeline --- landmark extraction, Fr\'echet mean, tangent-space embedding, geodesic regression, regional decomposition, classification --- applied to corpus callosum shape data from OASIS-1.

    \item We derive \emph{multivariate geodesic regression} with age, sex, and diagnosis as joint covariates, using the exact Riemannian gradient obtained via closed-form Jacobi fields of $\CP^{k-2}$ rather than a parallel-transport approximation.

    \item We quantify regional age-related shape change through a \emph{kinetic energy} decomposition of the fitted age-velocity field. Since Procrustes normalization has removed size, rotation, and translation, this energy reflects relative geometric deformation. The ranking is anterior-dominant: Genu $46.1\%$, Anterior Body $23.3\%$, Splenium $12.6\%$, Posterior Body $12.2\%$, Isthmus $5.9\%$ ($200$-sample bootstrap confirms the ordering).

    \item Signed-projection correlations confirm that the Genu tracks age most tightly ($r = 0.562$, $R^2 = 0.316$), followed by the Anterior Body ($r = 0.507$, $R^2 = 0.257$). Leave-one-region-out analysis identifies the Genu and Splenium as the only regions whose removal degrades the model ($\Delta R^2 = +0.0304$ and $+0.0175$), establishing the Genu as the dominant locus of size-normalized age-related shape variation. This convergence across energy, correlation, and leave-one-out is the paper's central empirical result.

    \item We correct a subtle unfairness in the standard Euclidean benchmark --- an unconstrained OLS intercept converges to a point that is not a valid shape --- and show that geodesic regression matches or outperforms the corrected benchmark, with the largest gap in the CN-only cohort ($\Delta R^2 \approx +0.014$).

    \item We benchmark against the volumetric classifier of \citep{soysal2025machine}: using only $128$ boundary points on the 2D midsagittal contour, our tangent-space representation achieves age-group classification AUC $0.791$, compared to their $0.67$ from a full 3D volumetric segmentation.
\end{itemize}

\subsection{Paper Organization}

Section~2 reviews the differential-geometric background required for the rest of the paper: Riemannian manifolds, tangent spaces, affine connections, and geodesics. Section~3 develops geodesic regression, contrasts it with linear regression, and generalizes it to the multivariate case. Section~4 introduces Kendall's shape space in detail, including the pre-shape sphere, the group action that is quotiented out, and the resulting quotient geometry. Section~5 describes the algorithms used in our pipeline (gradient descent, Fr\'echet mean, geodesic regression, regional energy decomposition) and reports the three experiments: nested regression across cohorts, young-vs-adult classification, and regional age-related shape change. Section~6 concludes with limitations and directions for future work.

\section{Mathematical Background}

This section develops the tools from differential geometry needed to make the rest of the paper self-contained. We follow the standard treatments in \citep{docarmo1992riemannian} and \citep{lee2018introduction}, and adopt the specific formulation of geodesic regression on Riemannian manifolds due to \citep{fletcher2013geodesic-regression}. The main constructions we use — Riemannian manifolds, tangent spaces, affine connections, geodesics, and the exponential and logarithm maps — are all standard, and we state them in the form most convenient for the Kendall shape-space computations that follow.

\subsection{Riemannian Manifolds}

\begin{definition}[Riemannian manifold]
A \emph{Riemannian manifold} is a pair $(\M, g)$ where $\M$ is a smooth $n$-dimensional manifold and $g$ is a \emph{Riemannian metric}: a smooth assignment $p \mapsto g_p$ of a positive-definite inner product $g_p : \T_p\M \times \T_p\M \to \R$ on the tangent space $\T_p\M$ at each point $p \in \M$.
\end{definition}

\begin{remark}
   The metric $g$ endows $\M$ with the local geometric notions of length, angle, and volume. For $v, w \in \T_p\M$ we write $\inner{v}{w}_p := g_p(v,w)$ and $\norm{v}_p := \sqrt{\inner{v}{v}_p}$.
\end{remark}

\begin{definition}[Tangent space]
At each point $p \in \M$, the \emph{tangent space} $\T_p\M$ is the vector space of equivalence classes of curves $\gamma : (-\varepsilon,\varepsilon) \to \M$ with $\gamma(0)=p$, two curves being equivalent if they have the same velocity vector $\gamma'(0)$ under any (equivalently, every) coordinate chart. Elements of $\T_p\M$ are tangent vectors at $p$; $\T_p\M$ is linear even though $\M$ itself is generally curved.
\end{definition}

Intuitively, $\T_p\M$ is the best linear (first-order) approximation of $\M$ at $p$ --- the ``flat'' space, one obtains by zooming in infinitely close to $p$. This linearization of a curved space enables us to apply statistical analysis tools (like linear regression, PCA, etc) on $\T_p\M$, an actual vector space, rather than directly on the curved space $\M$. 

\begin{definition}[Length and Riemannian distance]
The length of a piecewise-smooth curve $\gamma:[0,1]\to\M$ is
\[
L(\gamma) = \int_0^1 \norm{\gamma'(t)}_{\gamma(t)}\, dt .
\]
The \emph{Riemannian distance} between $p, q \in \M$ is
\[
\dist(p,q) = \inf\{ L(\gamma) : \gamma(0)=p,\ \gamma(1)=q \}.
\]
\end{definition}

\subsection{Affine Connections and Covariant Differentiation}

On a general manifold there is no canonical way to compare tangent vectors living in different tangent spaces $\T_p\M$ and $\T_q\M$; an \emph{affine connection} resolves this, and is the structure needed to differentiate vector fields and to define the notion of ``straight lines'' (geodesics) intrinsically on a manifold.

\begin{definition}[Affine connection]
An \emph{affine connection} $\nabla$ on $\M$ is a map
\[
\nabla : \mathfrak{X}(\M) \times \mathfrak{X}(\M) \to \mathfrak{X}(\M), \qquad (X,Y) \mapsto \nabla_X Y,
\]
where $\mathfrak{X}(\M)$ denotes smooth vector fields on $\M$, satisfying, for smooth functions $f,g$ and vector fields $X,Y,Z$:
\begin{align}
\nabla_{fX + gY} Z &= f\,\nabla_X Z + g\,\nabla_Y Z, \label{eq:conn1}\\
\nabla_X (Y+Z) &= \nabla_X Y + \nabla_X Z, \label{eq:conn2}\\
\nabla_X (fY) &= f\, \nabla_X Y + (Xf)\,Y. \label{eq:conn3}
\end{align}
\end{definition}

\begin{definition}
Let $V$ be a vector field along a differentiable curve $\gamma: I \rightarrow \M$. Suppose $V$ is induced by a vector field $Y \in \mathfrak{X}(M)$. That is, $V(t) = Y(\gamma(t))$. The \emph{Covariant Derivative} of $V$ along $c$ is given by $\frac{DV}{dt} = \nabla_{\frac{d\gamma}{dt}} Y$.
\end{definition}

The vector field $\nabla_X Y$ is called the covariant derivative of $Y$ along $X$. This can informally be interpreted as measuring the rate of change of $Y$ by an observer moving along the flow of $X$. 

Choosing a local coordinate system $(x_1, x_2, \ldots, x_n)$ at the point $p \in \M$, in this case, $X = \displaystyle \sum_i x_i X_i$ and $Y = \displaystyle \sum_j y_j X_j$ where $X_i = \frac{\partial}{\partial x_i}$, the covariant derivative is given by

\[\nabla_X Y = \sum_k \left(\sum_{ij} x_i y_j \Gamma^k_{ij} + X(y_k) \right) X_k \] where $\nabla_{X_i} X_j = \sum_k \Gamma_{ij}^k X_k$, the $\Gamma^k_{ij}$ are called Christoffel Symbols. In particular, given a curve $\gamma(t)$ and a vector field $V(t)$ defined along it. The covariant derivative of $V$ along $\gamma$ is given by:

\[\frac{DV^k}{dt} = \frac{dV^k}{dt} + \sum_{i,j} \Gamma_{ij}^k(\gamma(t))\, \dot\gamma^i(t)\, V^j(t) \]

%%%%%%%%%%%%%%%%%%%%%%%%%%%%%%%%%%
\subsection{Geodesics}

\begin{definition}[Geodesic]
A curve $\gamma : I \to \M$ is a \emph{geodesic} if its own velocity field is parallel along itself:
\[
\frac{D\dot\gamma}{dt} = 0, \qquad \text{i.e., in coordinates,} \qquad
\ddot\gamma^k(t) + \sum_{i,j} \Gamma_{ij}^k(\gamma(t))\, \dot\gamma^i(t)\, \dot\gamma^j(t) = 0, \quad k=1,\dots,n.
\]
\end{definition}

Geodesics generalize the notion of ``straight line'' in Eucldean spaces to  to curved spaces: they are curves of zero intrinsic acceleration, and (locally) they realize the shortest path between two points, i.e. $L(\gamma) = \dist(\gamma(0),\gamma(1))$ for $t$ in a sufficiently small interval. By the theory of ODEs, for every $p\in\M$ and $v\in\T_p\M$ there is a unique geodesic $\gamma_{p,v}(t)$ with $\gamma_{p,v}(0)=p$ and $\dot\gamma_{p,v}(0)=v$, at least for $t$ in a neighborhood of $0$.

%%%%%%%%%%%%%%%%%%%%%%%%%%%%%%%%%%%
\begin{definition}[Exponential and logarithm maps]
The \emph{Riemannian exponential map} at $p$ is
\[
\Exp_p : \T_p\M \to \M, \qquad \Exp_p(v) = \gamma_{p,v}(1),
\]
defined on a neighborhood of $0 \in \T_p\M$ where $\gamma_{p,v}(t)$ is the unique geodesic satisfying $\gamma_{p,v}(0) = p$ and $\dot{\gamma}_{p,v}(0) =v$. When $\Exp_p$ is a diffeomorphism from a neighborhood $U\subset \T_p\M$ of $0$ onto a neighborhood $V\subset\M$ of $p$, its inverse
\[
\Log_p : V \to U, \qquad \Log_p(q) = v \ \text{such that}\ \Exp_p(v) = q,\ \norm{v}_p = \dist(p,q),
\]
is called the \emph{Riemannian logarithm map}.
\end{definition}

%%%%%%%%%%%%%%%%%%%%%%%%%%%%%%%%%%%
The maps  $\Exp_p, \Log_p$ are the fundamental tools of this paper: $\Log_p$ \emph{linearizes} the manifold at $p$ by mapping nearby points $q \in \M$ to vectors $\Log_p(q) \in \T_p\M$ that point ``towards $q$'' and whose length equal to the true geodesic distance $\dist(p,q)$; $\Exp_p$ performs the inverse operation, mapping a tangent vector back onto the manifold. All of the standard statistical/geometrical tools developed below --- Fr\'echet means, PCA, and geodesic regression --- are built from repeated, careful use of this Exp/Log pair.

\begin{definition}[Fr\'echet mean]
Given points $p_1,\dots,p_N \in \M$, the (sample) \emph{Fr\'echet mean}, also called the \emph{Karcher mean}, is
\[
\bar p = \argmin_{p \in \M} \sum_{i=1}^N \dist(p, p_i)^2.
\]
\end{definition}

Unlike the Euclidean mean, $\bar p$ need not be unique on a general manifold (uniqueness is guaranteed locally under a bound on the sectional curvature and the diameter of the data, e.g., by results of Karcher \citep{karcher1977riemannian}), and it is in general be computed \emph{iteratively}, using gradient descent with the gradient of the sum-of-squared-distances functional given by
\[
\nabla_p \left[ \sum_i \dist(p,p_i)^2 \right] = -2\sum_i \Log_p(p_i).
\]
This identity is the basis of Algorithm~\ref{alg:frechet} in Section~5.
%%%%%%%%%%%%%%%%%%%%%%%%%%%%%%%%

\begin{definition}[Parallel transport]
Given a curve $\gamma:[0,1]\to\M$ with $\gamma(0)=p$, a vector field $V(t)$ along $\gamma$ is \emph{parallel} if $\tfrac{DV}{dt}=0$ for all $t$, i.e., it varies only by whatever change is forced by the curvature of $\M$, with no additional drift. For $v\in\T_p\M$, the \emph{parallel transport} $\Pi_{p\to q}(v) \in \T_q\M$ (with $q=\gamma(1)$) is the value $V(1)$ of the unique parallel vector field along $\gamma$ with $V(0)=v$. Parallel transport is a linear isometry $\T_p\M\to \T_q\M$: it preserves lengths and angles, and is the correct way to compare or combine tangent vectors that live at different points of a curved manifold.
\end{definition}

\begin{definition}[Jacobi field]
A \emph{Jacobi field} along a geodesic $\gamma$ is a vector field $J(t)$ satisfying the \emph{Jacobi equation}
\[
\frac{D^2 J}{dt^2} + R\big(J(t),\dot\gamma(t)\big)\dot\gamma(t) = 0,
\]
where $R$ is the Riemannian curvature tensor. Jacobi fields describe the infinitesimal separation between $\gamma$ and a neighboring geodesic, and are exactly the tool needed to differentiate the exponential map: for fixed $p,v$, $J(t) = D\Exp_p(tv)[t w]$ is the Jacobi field along $\gamma_{p,v}$ determined by the initial data $w\in\T_p\M$ (with boundary conditions depending on which argument of $\Exp$ is being perturbed). On a general manifold this ODE has no closed form; on a manifold of constant or piecewise-constant sectional curvature it reduces to a simple, explicit scalar equation, which is exactly the situation for Kendall shape space (Section~\ref{sec:jacobi}).
\end{definition}

%%%%%%%%%%%%%%%%%%%%%%%%%%%%%%%%%%%
\section{Geodesic Regression}

Geodesic regression generalizes linear regression to Riemannian manifolds,
replacing the fitted line with a geodesic. We develop the model, contrast it
with its Euclidean counterpart, and extend it to the multivariate case,
following \citep{fletcher2013geodesic-regression}.

%==================================================================

\subsection{From Linear Regression to Geodesic Regression}

We begin with ordinary least squares, whose structure we generalize in the
following section.

\subsubsection{Linear regression (review)}

In ordinary least-squares linear regression, we observe pairs $(t_i, y_i) \in \R \times \R^m$, $i=1,\dots,N$, and we assume a model
\[
y_i = \alpha + \beta\, t_i + \epsilon_i, \qquad \alpha \in \R^m,\ \beta \in \R^m,\ \epsilon_i \sim \mathcal{N}(0,\sigma^2 I),
\]
and estimate $(\alpha,\beta)$ by minimizing the residual sum of squares $\displaystyle \sum_i \norm{y_i - \alpha - \beta t_i}^2$. Both the response $y_i$ and the fitted curve $t \mapsto \alpha + \beta t$ live in the flat vector space $\R^m$, so ``$+$'' and ``$-$'' are unambiguous, and the objective is a smooth, convex quadratic in $(\alpha,\beta)$ admitting a closed-form (normal-equations) solution.

%%%%%%%%%%%%%%%%%%%%%%%%%%%%%%%%%%%%
\subsubsection{Geodesic regression}

When the target $y_i$ instead lies on a curved Riemannian manifold $\M$ (in our case, $\M$ is Kendall's shape space- that will be defined below and $y_i$ is the shape of subject $i$'s corpus callosum), the additive model from linear regression is no longer well-defined, because $\M$ is not a vector space and ``$\alpha + \beta t$'' has no intrinsic meaning in $\M$. \citep{fletcher2013geodesic-regression} introduced \emph{geodesic regression} to resolve this: the regression line $\alpha + \beta t$ is replaced by a \emph{geodesic curve} $\gamma(t) = \Exp_{p_0}(t\, v)$ parametrized by a base point $p_0\in\M$ (the geometric analogue of the intercept $\alpha$) and a tangent vector $v \in \T_{p_0}\M$ (the analogue of the slope $\beta$). Given observations $(t_i, y_i) \in \R \times \M$, we minimize the sum of squared \emph{geodesic (Riemannian) distances}:
\begin{equation}
(p_0^\ast, v^\ast) = \argmin_{p_0 \in \M,\ v \in \T_{p_0}\M} \ \sum_{i=1}^N \dist\!\big(\Exp_{p_0}(t_i v),\, y_i\big)^2.
\label{eq:geodreg}
\end{equation}
%%%%%%%%%%%%%%%%%%%%%%%%%%%%%%%%%%%%%
Table~\ref{tab:lin-vs-geo} summarizes the correspondence.

\begin{table}[H]
\centering
\caption{Correspondence between linear regression and geodesic regression.}
\label{tab:lin-vs-geo}
\begin{tabular}{@{}lll@{}}
\toprule
\textbf{Concept} & \textbf{Linear regression} ($\R^m$) & \textbf{Geodesic regression} ($\M$) \\
\midrule
Ambient space & Vector space $\R^m$ & Riemannian manifold $\M$ \\
Model curve & $t \mapsto \alpha + \beta t$ & $t \mapsto \Exp_{p_0}(t\,v)$ \\
Intercept & $\alpha \in \R^m$ & Base point $p_0 \in \M$ \\
Slope & $\beta \in \R^m$ & Tangent vector $v \in \T_{p_0}\M$ \\
Residual & $y_i - (\alpha + \beta t_i)$ & $\Log_{\Exp_{p_0}(t_i v)}(y_i)$ \\
Loss & $\sum_i \norm{y_i - \alpha - \beta t_i}^2$ & $\sum_i \dist(\Exp_{p_0}(t_i v), y_i)^2$ \\
Solution method & Closed form (normal equations) & Iterative (Riemannian gradient descent) \\
\bottomrule
\end{tabular}
\end{table}

Two structural differences are worth mentioning. First, because $\M$ is curved, the objective in \eqref{eq:geodreg} is generally \emph{non-convex}, so unlike ordinary least squares there is no guarantee that gradient-based optimization converges to the global optimum; this motivates our Experiment~1 in Section~5, which studies sensitivity to initialization. Second, both the ``intercept'' $p_0$ and the ``slope'' $v$ must be updated jointly and consistently — since $v$ lives in $\T_{p_0}\M$, moving $p_0$ changes the very vector space in which $v$ is expressed. This coupling is of no concern in the Euclidean case and on general smooth manifold, it is handled using parallel transport (Section~3.3).
%%%%%%%%%%%%%%%%%%%%%%%%%%%%%%%%%%%%%%
\subsection{Multivariate Geodesic Regression}

\subsubsection{Multiple linear regression (review)}

In multiple linear regression, the scalar covariate $t_i$ is replaced by a covariate vector $\mathbf{x}_i = (x_{i1}, \dots, x_{iK}) \in \R^K$ (e.g., age, sex, education), and the model becomes
\[
y_i = \alpha + \sum_{k=1}^K \beta_k x_{ik} + \epsilon_i = \alpha + B \mathbf{x}_i + \epsilon_i,
\]
where $B \in \R^{m \times K}$ collects the $K$ slope vectors $\beta_1,\dots,\beta_K \in \R^m$ as its columns. Each $\beta_k$ measures the marginal effect on $y$ of the $k$-th covariate, holding the others fixed.

%%%%%%%%%%%%%%%%%%%%%%%%%%%%%%%%%%%%%%
\subsubsection{Multivariate geodesic regression}

We now present the multivariate extension of geodesic regression, formulating it as a natural generalization of Fletcher's single-covariate model \citep{fletcher2013geodesic-regression} in the exact same manner that multiple linear regression extends simple linear regression. Let $\mathbf{x}_i = (x_{i1},\dots,x_{iK}) \in \R^K$ be the covariates of subject $i$ (e.g., $x_{i1}=$ age, $x_{i2}=$ sex indicator, $x_{i3}=$ diagnosis indicator), and let $y_i \in \M$ be the corresponding shape. We posit a base point $p_0 \in \M$ and $K$ tangent directions $v_1,\dots,v_K \in \T_{p_0}\M$, one per covariate, and define the model curve/surface
\begin{equation}
\hat y(\mathbf{x}) \;=\; \Exp_{p_0}\!\left( \sum_{k=1}^K x_k\, v_k \right).
\label{eq:multivar-geo}
\end{equation}
Because $\T_{p_0}\M$ is a genuine (linear) vector space, the sum $\sum_k x_k v_k$ inside the exponential map is an ordinary linear combination — this is precisely where the manifold structure is exploited: \emph{all} of the ``multivariate'' linear combination happens safely inside the flat tangent space, and only the final result is mapped back onto the curved manifold via a single application of $\Exp_{p_0}$. The parameters $(p_0, v_1,\dots,v_K)$ are estimated by minimizing
\begin{equation}
\mathcal{L}(p_0, v_1,\dots,v_K) \;=\; \sum_{i=1}^N \dist\!\Big(\Exp_{p_0}\big(\textstyle\sum_{k} x_{ik} v_k\big),\, y_i\Big)^2 .
\label{eq:multivar-loss}
\end{equation}
Setting $K=1$ and $x_{i1}=t_i$ recovers the single-covariate model \eqref{eq:geodreg}, exactly as simple linear regression is the $K=1$ case of multiple linear regression. Each fitted $v_k$ admits the same interpretation as $\beta_k$ in the Euclidean case: it is the (geodesic) direction and rate of shape change associated with a unit increase in covariate $k$, holding the other covariates fixed, expressed in the tangent space at the common base point $p_0$.

\begin{remark}[Coefficient of determination on manifolds]
By analogy with the Euclidean $R^2$, \citet{fletcher2013geodesic-regression} defines a Riemannian coefficient of determination
\[
R^2 = 1 - \frac{\sum_i \dist(\hat y(\mathbf{x}_i), y_i)^2}{\sum_i \dist(\bar y, y_i)^2},
\]
where $\bar y$ is the Fr\'echet mean of $\{y_i\}$ (the model with $v_1=\dots=v_K=0$). We report this quantity for our fitted models in Section~5.
\end{remark}
%%%%%%%%%%%%%%%%%%%%%%%%%%%%%%
\subsection{Exact Riemannian Gradients via the Differential of the Exponential Map}
\label{sec:exact-gradient}

Minimizing \eqref{eq:multivar-loss} by gradient descent requires the true Riemannian gradients $\nabla_{p_0}\mathcal L$ and $\nabla_{v_k}\mathcal L$, and this is where the curvature of $\M$ enters the \emph{computation itself}, not just the model. Write $u_i = \sum_k x_{ik} v_k \in \T_{p_0}\M$ for the combined tangent direction of subject $i$, $\hat y_i = \Exp_{p_0}(u_i)$ for the predicted shape, and
\[
\epsilon_i \;=\; \Log_{\hat y_i}(y_i) \;\in\; \T_{\hat y_i}\M
\]
for the residual tangent vector, pointing from the prediction $\hat y_i$ towards the observation $y_i$. Since $\mathcal L = \sum_i \dist(\hat y_i, y_i)^2 = \sum_i \norm{\epsilon_i}^2$, the chain rule requires differentiating $\hat y_i = \Exp_{p_0}(u_i)$ with respect to \emph{both} arguments of $\Exp$. Write $D_1\Exp_{(p,v)} : \T_p\M \to \T_{\Exp_p(v)}\M$ for the differential of $\Exp$ with respect to its base point (holding the tangent vector fixed), and $D_2\Exp_{(p,v)} : \T_p\M \to \T_{\Exp_p(v)}\M$ for the differential with respect to the tangent vector (holding the base point fixed). Fletcher's derivation of the gradient of a sum-of-squared-distances objective along a geodesic \citep{fletcher2013geodesic-regression} gives
\begin{align}
\nabla_{p_0}\mathcal L &= -2\sum_{i=1}^N D_1\Exp_{(p_0,u_i)}^{\ast}(\epsilon_i), \label{eq:grad-p}\\
\nabla_{v_k}\mathcal L &= -2\sum_{i=1}^N x_{ik}\, D_2\Exp_{(p_0,u_i)}^{\ast}(\epsilon_i), \qquad k=1,\dots,K, \label{eq:grad-v}
\end{align}
where $D_1\Exp^\ast, D_2\Exp^\ast : \T_{\hat y_i}\M \to \T_{p_0}\M$ denote the adjoints of $D_1\Exp$ and $D_2\Exp$ with respect to the Riemannian metric — the operators that correctly ``pull back'' a perturbation of the output $\hat y_i$ to the corresponding perturbation of the input $(p_0, u_i)$.

%%%%%%%%%%%%%%%%%%%%%%%%%%%%%%
\subsection{Gradient Descent for Geodesic Regression: Pseudocode}

Algorithm~\ref{alg:georeg} gives Riemannian gradient descent for the multivariate model \eqref{eq:multivar-loss}; setting $K=1$ recovers the single-covariate case \eqref{eq:geodreg}. The per-sample gradient contributions are obtained via the chain rule applied to $\dist(\cdot,\cdot)^2$ together with the differential of $\Exp_{p_0}$ (see \cite{fletcher2013geodesic-regression} for the full derivation on general manifolds; in Kendall shape space these quantities have closed forms, given in Section~4.4).

\begin{algorithm}[H]
\caption{Riemannian Gradient Descent for Multivariate Geodesic Regression (Exact Gradients)}
\label{alg:georeg}
\begin{algorithmic}[1]
\Require Data $\{(\mathbf{x}_i, y_i)\}_{i=1}^N$, $\mathbf{x}_i \in \R^K$, $y_i \in \M$; learning rate $\eta$; max iterations $T$; tolerance $\tau$
\Ensure Estimated base point $p_0 \in \M$ and slopes $v_1,\dots,v_K \in \T_{p_0}\M$
\State Initialize $p_0^{(0)} \gets$ Fr\'echet mean of $\{y_i\}$ \Comment{see Algorithm~\ref{alg:frechet}}
\State Initialize $v_k^{(0)} \gets 0 \in \T_{p_0^{(0)}}\M$ for $k=1,\dots,K$
\For{$t = 0, 1, \dots, T-1$}
    \For{$i = 1,\dots,N$}
        \State $u_i \gets \sum_{k=1}^K x_{ik}\, v_k^{(t)}$ \Comment{combined tangent direction}
        \State $\hat y_i \gets \Exp_{p_0^{(t)}}(u_i)$ \Comment{predicted shape}
        \State $\epsilon_i \gets \Log_{\hat y_i}(y_i)$ \Comment{residual tangent vector at $\hat y_i$}
        \State $g_{p_0,i} \gets D_1\Exp_{(p_0^{(t)},\,u_i)}^{\ast}(\epsilon_i)$ \Comment{exact adjoint, Eq.~\eqref{eq:jacobi-point} (cosine solution)}
        \State $g_{v,i} \gets D_2\Exp_{(p_0^{(t)},\,u_i)}^{\ast}(\epsilon_i)$ \Comment{exact adjoint, Eq.~\eqref{eq:jacobi-velocity} (sine solution)}
    \EndFor
    \State $g_{p_0} \gets -\frac{2}{N}\sum_{i=1}^N g_{p_0,i}$ \Comment{$=\nabla_{p_0}\mathcal L$, Eq.~\eqref{eq:grad-p}}
    \State $g_{v_k} \gets -\frac{2}{N}\sum_{i=1}^N x_{ik}\, g_{v,i}$ \quad for $k=1,\dots,K$ \Comment{$=\nabla_{v_k}\mathcal L$, Eq.~\eqref{eq:grad-v}}
    \State $p_0^{(t+1)} \gets \Exp_{p_0^{(t)}}(-\eta\, g_{p_0})$ \Comment{move base point along the manifold}
    \For{$k = 1,\dots,K$}
        \State $v_k^{(t+1)} \gets \Pi_{p_0^{(t)} \to p_0^{(t+1)}}\!\big(v_k^{(t)} - \eta\, g_{v_k}\big)$ \Comment{update slope \& re-express at new base point}
    \EndFor
    \If{$\big|\mathcal{L}^{(t+1)} - \mathcal{L}^{(t)}\big| < \tau$}
        \State \textbf{break}
    \EndIf
\EndFor
\State \Return $p_0^{(T)}, v_1^{(T)}, \dots, v_K^{(T)}$
\end{algorithmic}
\end{algorithm}

\begin{remark}
When $\M$ is Kendall shape space $\CP^{k-2}$ (Section~4), the maps $\Exp$, $\Log$, $\Pi$, and the exact adjoint operators $D_1\Exp^\ast, D_2\Exp^\ast$ all have closed forms (Eqs.~\eqref{eq:cp-exp}--\eqref{eq:cp-parallel} and \eqref{eq:jacobi-point}--\eqref{eq:jacobi-velocity}), so every line of Algorithm~\ref{alg:georeg} can be evaluated exactly, without any inner optimization loop.
\end{remark}

%%%%%%%%%%%%%%%%%%%%%%%%%%%%%%%%%%%%%

%==================================================================
\section{Kendall Shape Space}
%==================================================================

\subsection{Landmark Configurations and the Pre-shape Sphere}

Following \citep{kendall1984shape} and the book \citep{dryden2016statistical}, we represent an anatomical contour by $k$ ordered, labeled landmark points in the plane, collected into a configuration matrix
\[
X = (x_1, \dots, x_k)^\top \in \R^{k\times 2}, \qquad x_j = (x_j^{(1)}, x_j^{(2)}) \in \R^2 .
\]
Identifying $\R^2 \cong \C$, we equivalently write $X \in \C^k$ with $j$-th entry $z_j = x_j^{(1)} + i\, x_j^{(2)}$.

\paragraph{Step 1:  Remove translation.} Subtract the centroid $\bar z = \frac{1}{k}\sum_j z_j$ from every landmark:
\[
\tilde z = z - \bar z \mathbf{1}_k .
\]
This projects $X$ onto the centered subspace $\C^k_0 = \{ z \in \C^k : \sum_j z_j = 0\} \cong \C^{k-1}$.

\paragraph{Step 2:Remove scale.} Divide by the centroid size $\norm{\tilde z} = \big(\sum_j |\tilde z_j|^2\big)^{1/2}$:
\[
w = \tilde z / \norm{\tilde z} \in \C^{k-1}, \qquad \norm{w}=1 .
\]
The resulting $w$ lies on the unit sphere $S^{2k-3} \subset \C^{k-1} \cong \R^{2k-2}$, called the \emph{pre-shape sphere} $\preshape_2^k$.

\paragraph{Step 3: Remove rotation.} A planar rotation by angle $\theta$ acts on $w$ by complex multiplication, $w \mapsto e^{i\theta} w$. Kendall's shape space is defined as the quotient of the pre-shape sphere by this residual $S^1 = U(1)$ action:
\[
\Sigma_2^k \;:=\; \preshape_2^k / \mathrm{SO}(2) \;=\; S^{2k-3}\Big / (e^{i\theta}w \sim w, \theta \in [0, 2 \pi) ).
\]

\begin{theorem}[Kendall]
\label{thm:kendall}
For planar landmarks ($m=2$), the shape space $\Sigma_2^k$ is diffeomorphic to complex projective space $\CP^{k-2}$, with the quotient (Fubini--Study) metric induced by the Hopf fibration $S^{2k-3} \to \CP^{k-2}$.
\end{theorem}

For $k=128$ landmarks, $\Sigma_2^{128} \cong \CP^{126}$, a manifold of complex dimension $126$ and real dimension $252$, matching the value stated in the abstract.

\subsection{Riemannian Metric on $\CP^{k-2}$}

Points of $\CP^{k-2}$ are equivalence classes $[w] = \{e^{i\theta}w : \theta \in [0,2\pi)\}$ of unit vectors $w \in \C^{k-1}$. The Riemannian (geodesic) distance between two shapes $[w_1], [w_2]$ is
\begin{equation}
\dist\big([w_1],[w_2]\big) \;=\; \arccos\big| \inner{w_1}{w_2}_{\C} \big|,
\label{eq:cp-dist}
\end{equation}
where $\inner{w_1}{w_2}_\C = \sum_j \overline{w_{1,j}}\, w_{2,j}$ is the Hermitian inner product on $\C^{k-1}$, and $|\cdot|$ denotes the modulus of the resulting complex number. This is sometimes called the \emph{Procrustes distance}: it is precisely the optimal-rotation-aligned root-mean-square landmark distance between the two shapes. This quantity is also known as the \emph{full Procrustes distance} and can be equivalently expressed as
\[
d_F([w_1],[w_2]) = \sqrt{1 - |\inner{w_1}{w_2}_\C|^2},
\]
which agrees with $\sin(\dist)$ for the geodesic distance in \eqref{eq:cp-dist} at leading order and is often used interchangeably for small shape differences.

\subsection{Tangent Space, Exponential and Logarithm Maps}

Fix a base shape $[p] \in \CP^{k-2}$ represented by a unit vector $p \in \C^{k-1}$. The (horizontal) tangent space at $[p]$ may be identified with
\[
\T_{[p]}\CP^{k-2} \;\cong\; \{ v \in \C^{k-1} : \inner{p}{v}_\C = 0 \},
\]
i.e., the orthogonal complement of $p$ (both to the real direction $p$ itself and to the ``vertical'' direction $ip$ generated by the rotation action, once $v$ is required to be Hermitian-orthogonal to $p$). For $v \in \T_{[p]}\CP^{k-2}$ with $\theta = \norm{v} \neq 0$, the exponential map is
\begin{equation}
\Exp_{[p]}(v) \;=\; \Big[ \cos(\theta)\, p \;+\; \sin(\theta) \frac{v}{\theta} \Big],
\label{eq:cp-exp}
\end{equation}
and, given a second shape $[q]$ represented by a unit vector $q$ with $\inner{p}{q}_\C \neq 0$, first rotate $q$ into optimal alignment with $p$,
\[
q^\ast = q\, \frac{\overline{\inner{p}{q}_\C}}{|\inner{p}{q}_\C|},
\]
(this is the Procrustes alignment step, undoing the residual rotational freedom) so that $\inner{p}{q^\ast}_\C \in \R_{>0}$; then the logarithm map is
\begin{equation}
\Log_{[p]}([q]) \;=\; \frac{\theta}{\sin\theta}\big( q^\ast - \cos(\theta)\, p \big), \qquad \theta = \arccos\big(\mathrm{Re}\,\inner{p}{q^\ast}_\C\big) = \dist([p],[q]).
\label{eq:cp-log}
\end{equation}
Parallel transport of a tangent vector $v \in \T_{[p]}\CP^{k-2}$ along the geodesic from $[p]$ towards $[q]$, to the point reached at arclength $t$, is given in closed form by
\begin{equation}
\Pi_{[p]\to \gamma(t)}(v) \;=\; v - \frac{\inner{u}{v}}{\theta}\Big( \sin(t)\, p + \big(1-\cos(t)\big) u \Big), \qquad u = \Log_{[p]}([q]) / \theta,
\label{eq:cp-parallel}
\end{equation}
where $\theta = \dist([p],[q])$ and the inner product is the real part of the Hermitian pairing restricted to the tangent space. These closed forms (Eqs.~\eqref{eq:cp-exp}-\eqref{eq:cp-parallel}) are what make Algorithms~\ref{alg:frechet} and \ref{alg:georeg} directly and exactly implementable on Kendall shape space.
%%%%%%%%%%%%%%%%%%%%%%%%%%%%%%%%%%%%%%
\subsection{Curvature and Jacobi Fields on $\CP^{k-2}$}
\label{sec:jacobi}

Complex projective space is a Riemannian symmetric space of \emph{non-constant} sectional curvature: at any point $[p]$, along any geodesic direction $\hat v = v/\norm{v}$, the sectional curvature of the $2$-plane spanned by $\hat v$ and a unit vector $w\perp\hat v$ takes exactly one of two values \citep{docarmo1992riemannian, fletcher2013geodesic-regression}:
\begin{equation}
K(\hat v, w) =
\begin{cases}
4, & w \parallel J\hat v, \\[2pt]
1, & w \perp J\hat v,
\end{cases}
\label{eq:cp-curvature}
\end{equation}
where $J\hat v = i\hat v$ is the almost-complex structure on $\CP^{k-2}$ — multiplication of the ambient complex vector by $i$, which is exactly the infinitesimal generator of the residual $U(1)$ rotation quotiented out in Section~4.1, restricted to the horizontal tangent space. Geometrically, $\CP^{k-2}$ bends twice as sharply along the ``holomorphic'' direction $J\hat v$ as it does along every other direction orthogonal to the geodesic. 

\paragraph{Jacobi field decomposition.} Fix a geodesic $\gamma(t)=\Exp_p(tv)$ with arc length $L=\norm{v}$ (so $\gamma$ has constant speed $L$ on $t\in[0,1]$). Given $w\in\T_p\M$, decompose it relative to $\hat v = v/L$ as
\begin{equation}
w = w_\parallel + w_J + w_\perp, \qquad
w_\parallel = \inner{w}{\hat v}\hat v, \quad
w_J = \inner{w}{J\hat v}\,J\hat v, \quad
w_\perp = w - w_\parallel - w_J,
\label{eq:jacobi-decomp}
\end{equation}
the components along the geodesic direction itself (curvature-free, since $\dot\gamma$ is a trivial Jacobi field), along the holomorphic direction ($K=4$), and along everything else orthogonal to both ($K=1$). For a Jacobi field with $J(t)$ orthogonal to $\dot\gamma$ on a $2$-plane of constant curvature $K$, the Jacobi equation reduces to the scalar oscillator equation $J'' + KL^2 J = 0$; solving it with the two boundary conditions relevant to differentiating $\Exp_p(v)$ with respect to each of its two arguments gives:
\begin{align}
D_1\Exp_{(p,v)}^{\ast}(w) &\;=\; w_\parallel \;+\; \cos(L)\, w_\perp \;+\; \cos(2L)\, w_J,
\label{eq:jacobi-point}\\[4pt]
D_2\Exp_{(p,v)}^{\ast}(w) &\;=\; w_\parallel \;+\; \frac{\sin(L)}{L}\, w_\perp \;+\; \frac{\sin(2L)}{2L}\, w_J.
\label{eq:jacobi-velocity}
\end{align}
Eq.~\eqref{eq:jacobi-point} is the $J(0)=w,\,J'(0)=0$ (cosine) solution, appropriate for perturbing the geodesic's \emph{starting point} while holding its initial velocity fixed — exactly the differential $D_1\Exp$ needs. Eq.~\eqref{eq:jacobi-velocity} is the $J(0)=0,\,J'(0)=w$ (sine) solution, appropriate for perturbing the \emph{initial velocity} while holding the starting point fixed — exactly $D_2\Exp$. Because $\CP^{k-2}$ is a Riemannian symmetric space, both operators are self-adjoint \citep{fletcher2013geodesic-regression}, so \eqref{eq:jacobi-point}--\eqref{eq:jacobi-velocity} are simultaneously the forward differentials of $\Exp$ \emph{and} the adjoint operators $D_1\Exp^\ast, D_2\Exp^\ast$ required in Algorithm~\ref{alg:georeg}. In the degenerate case $L\to 0$ (no covariate effect), both operators reduce continuously to the identity, matching the flat-space limit.

%%%%%%%%%%%%%%%%%%%%%%%%%%%%%%%%%%%%%%%%
\subsection{From Corpus Callosum Contours to Points in $\CP^{126}$}\label{sec:cc-to-points}

For each subject, the midsagittal CC boundary is segmented from the T1-weighted MRI and resampled to $k=128$ equally spaced, correspondence-matched landmarks (using arc-length parametrization and a fixed anatomical starting landmark, e.g., the rostrum), yielding $X \in \R^{128\times 2}$. Centering, scaling, and Procrustes alignment (Steps 1-3 above) map $X$ to a point $[w] \in \CP^{126}$. Once every subject's shape has been embedded, the Fr\'echet mean $\overline{[w]}$ of the dataset is computed (Algorithm~\ref{alg:frechet}) and used as the base point for $\Log$; the resulting tangent vectors $\Log_{\overline{[w]}}([w_i]) \in \T_{\overline{[w]}}\CP^{126} \cong \R^{252}$ are the Euclidean-space representation on which PCA and the classifiers of Section~5 are built.

\begin{figure}[H]
    \centering
    
    % Top row: two images side by side
    \begin{subfigure}[b]{0.48\linewidth}
        \centering
        \includegraphics[width=\linewidth]{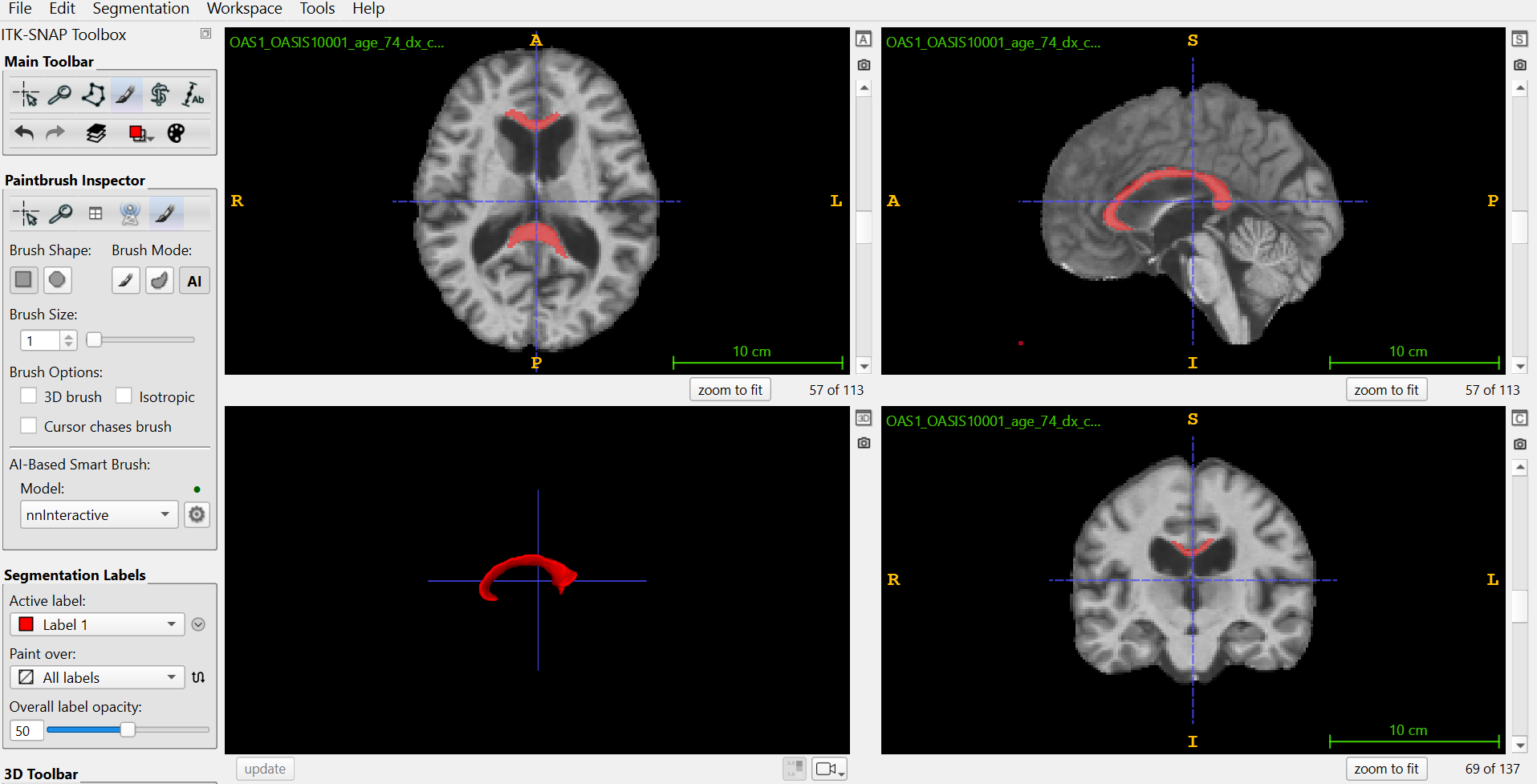}
        %\caption{Mean shape with landmarks colored by region.}
        \label{fig:mean_shape}
    \end{subfigure}
    \hfill
    \begin{subfigure}[b]{0.48\linewidth}
        \centering
        \includegraphics[width=\linewidth]{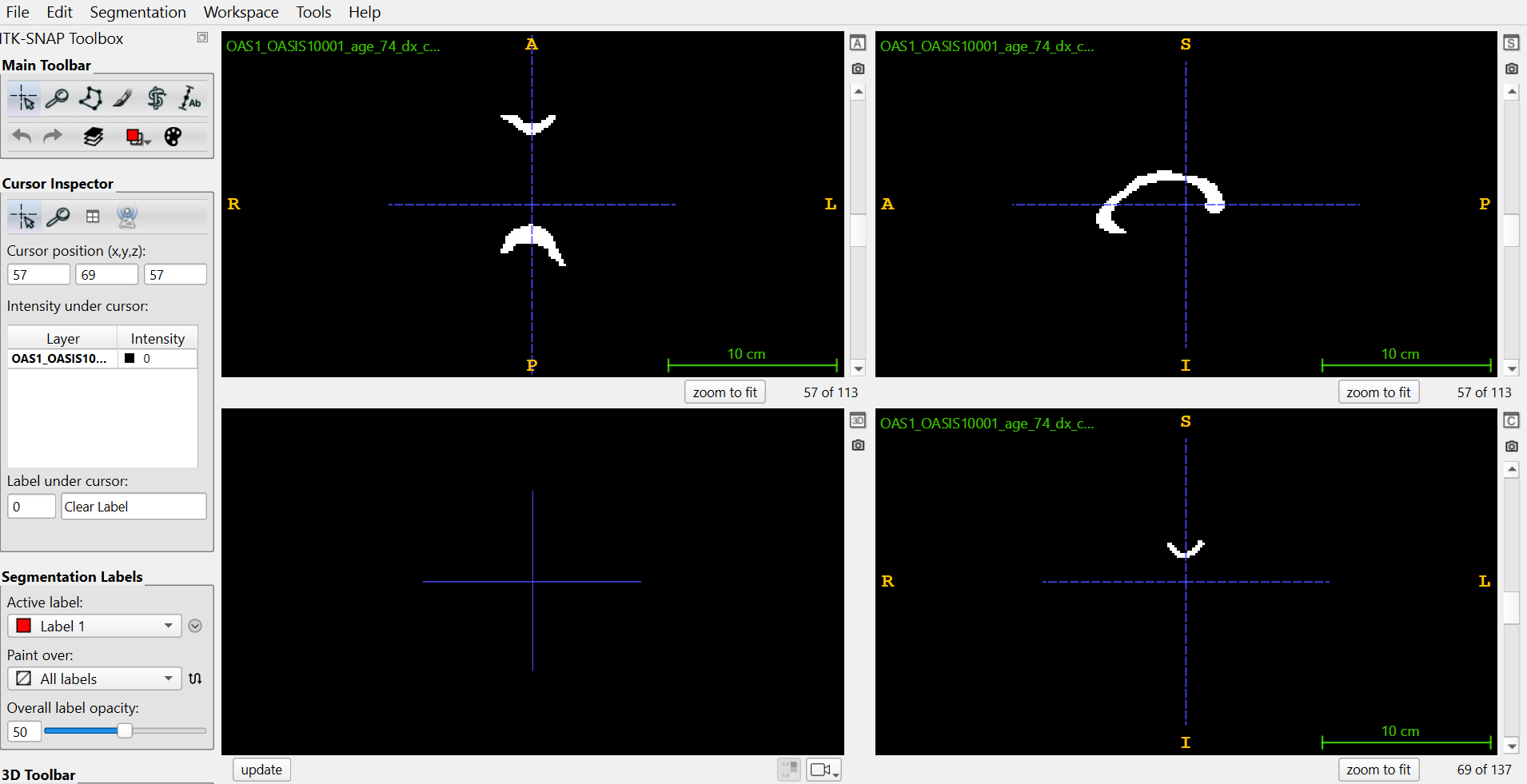}
        %\caption{Trajectory along the age velocity from $-2$ to $+2$ SD.}
        \label{fig:trajectory}
    \end{subfigure}
    
    \vspace{1em}
    
    % Bottom row: one image spanning the column
    \begin{subfigure}[b]{\linewidth}
        \centering
        \includegraphics[width=0.9\linewidth]{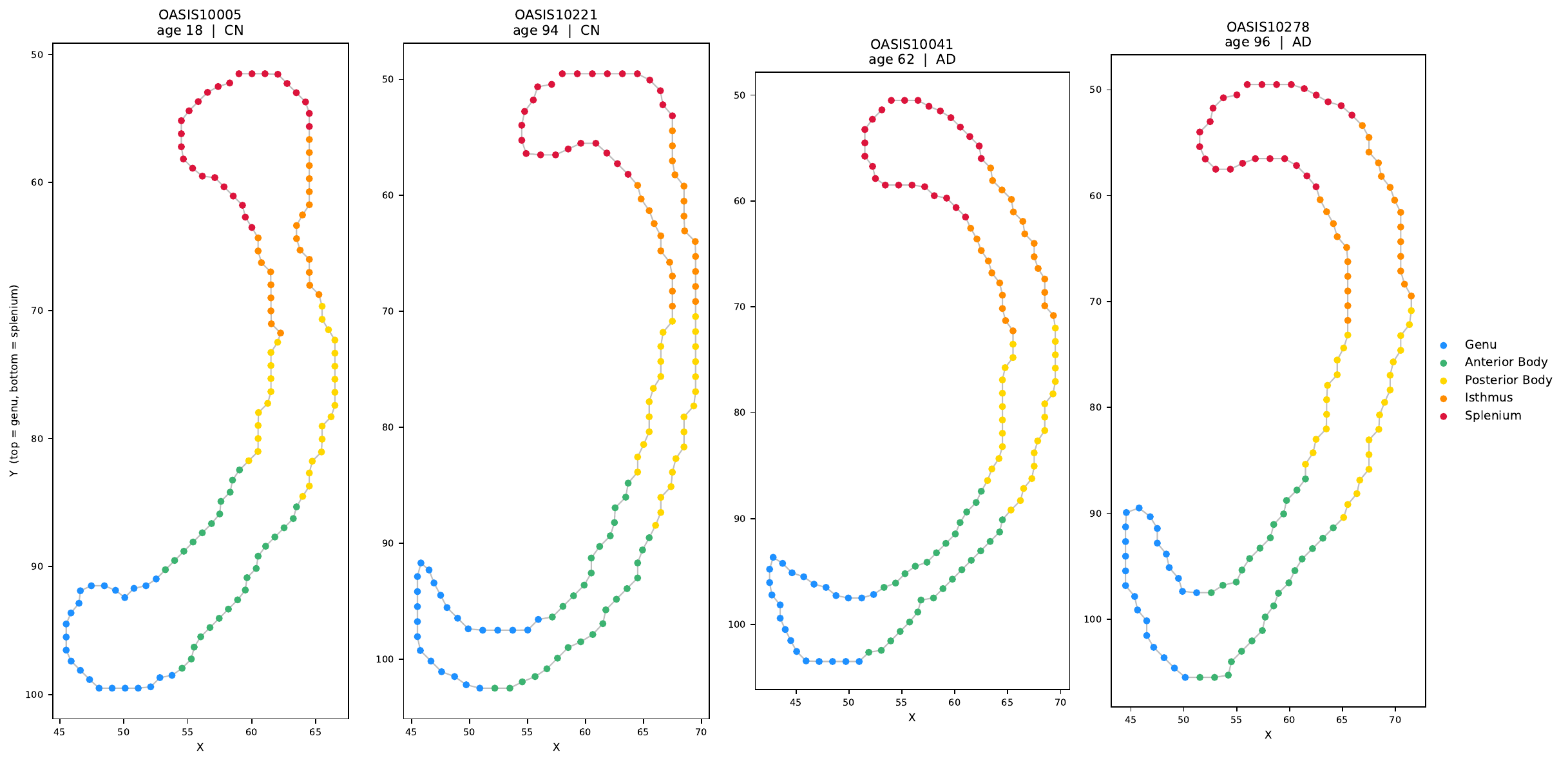}
        %\caption{Midsagittal segmentation of the corpus callosum with the extracted boundary overlaid.}
        \label{fig:segmentation}
    \end{subfigure}
    
    \caption{Midsagittal segmentation of the corpus callosum}
    \label{fig:composite}
\end{figure}

%==================================================================
\section{Algorithms and Experiments}
%==================================================================

\subsection{Fr\'echet Mean via Gradient Descent}

\begin{algorithm}[H]
\caption{Riemannian Gradient Descent for the Fr\'echet Mean}
\label{alg:frechet}
\begin{algorithmic}[1]
\Require Shapes $[w_1],\dots,[w_N] \in \M$; learning rate $\eta$; max iterations $T$; tolerance $\tau$
\Ensure Fr\'echet mean estimate $\bar{p}$
\State Initialize $\bar p^{(0)} \gets [w_{i_0}]$ for some (or random) index $i_0$ \Comment{initialization matters -- see Experiment 1}
\For{$t=0,1,\dots,T-1$}
    \For{$i=1,\dots,N$}
        \State $u_i \gets \Log_{\bar p^{(t)}}([w_i])$
    \EndFor
    \State $g \gets -\frac{2}{N}\sum_{i=1}^N u_i$ \Comment{gradient of $\sum_i d(\cdot,[w_i])^2$}
    \State $\bar p^{(t+1)} \gets \Exp_{\bar p^{(t)}}(-\eta\, g)$
    \State $F^{(t+1)} \gets \frac{1}{N}\sum_i \dist(\bar p^{(t+1)}, [w_i])^2$
    \If{$|F^{(t+1)} - F^{(t)}| < \tau$}
        \State \textbf{break}
    \EndIf
\EndFor
\State \Return $\bar p^{(T)}$
\end{algorithmic}
\end{algorithm}

\subsection{Experimental Setup}

\paragraph{Data.} We use $N=340$ subjects from the OASIS-1 cross-sectional MRI dataset \citep{marcus2007oasis}: $252$ cognitively normal (CN) participants and $88$ patients diagnosed with Alzheimer's disease (AD). For each subject, the midsagittal CC contour is segmented and represented with $k=128$ landmarks as described in Section~\ref{sec:cc-to-points}, and each subject's chronological age and gender (and, where used, diagnosis label) is recorded as covariate metadata. Two cohorts are analyzed: (i) the \emph{CN-only} cohort ($n=252$), isolating pure age-related shape change from confounding disease effects, and (ii) \emph{AD- only} cohort ($n=88$). 

%(iii) the \emph{complete cohort} ($n=340$, CN+AD), which additionally captures disease-related morphological change, which itself correlates with age.

\paragraph{Pipeline.} (1) Fit a multivariate geodesic regression with age and
gender as covariates for the CN and AD cohorts separately, and with age,
gender, and diagnosis as covariates for the pooled CN+AD cohort.
(2) Compute the Fr\'echet mean shape $\bar p$ of the training set via
Algorithm~\ref{alg:frechet}.
(3) Map every shape to the tangent space at $\bar p$ via \eqref{eq:cp-log},
yielding vectors in $\R^{252}$, whose first $128$ complex coordinates
correspond one-to-one with the $128$ contour landmarks.
(4) Apply PCA to the tangent vectors, retaining the top components explaining
the desired proportion of variance.
(5) Fit classifiers (Logistic Regression, Random Forest, Support Vector
Machine with RBF kernel) on the PCA scores to predict a binarized age-group
label (young vs.\ old, split at the cohort median or a clinically motivated
cutoff).
(6) Evaluate with stratified $k$-fold cross-validation, reporting AUC (area
under the ROC curve) as the primary metric, together with accuracy.
(7) For the regional analysis of Experiment~3, assign each of the $128$
landmarks to one of five anatomical sub-regions (Genu, Anterior Body,
Posterior Body, Isthmus, Splenium) using per-subject arc-length fractions
anchored at the genu and splenium tips, then take a cohort-level majority
vote across subjects to fix a single landmark-to-region map for the whole
analysis.
(8) Extract the fitted age-velocity tangent $v_\text{age} \in \T_{[p_0]}\CP^{126}$
from the single-covariate geodesic regression, and read off the per-landmark
squared magnitude $|v_\text{age}[j]|^2$ as a local kinetic energy.
(9) Aggregate these magnitudes within each region to obtain the regional
energy shares $E_R / \sum_{R'} E_{R'}$, and use a $200$-sample bootstrap
resampling of the CN cohort to attach confidence intervals to the regional
ranking.
(10) For each region $R$, compute the signed projection
$s_i^{(R)} = \mathrm{Re}\big(\sum_{j \in R} \overline{v_\text{age}[j]} \cdot w_i[j]\big)$
of every subject's tangent deviation $w_i = \Log_{[p_0]}([q_i])$ onto the
regional age direction, and correlate $s_i^{(R)}$ with chronological age to
measure how tightly each region tracks age.
(11) Perform a leave-one-region-out analysis by refitting the geodesic
regression on the reduced landmark set $\{1,\dots,128\} \setminus R$ and
recording the change in geodesic $R^2$, to identify regions whose removal
degrades the model.
(12) Fit each region's shape against age in isolation (\emph{solo fits}) to
contrast regions whose shape is predominantly age-determined against those
that merely carry a large share of the global age signal.

\paragraph{Baselines.} As a Euclidean baseline, the same classifiers are trained directly on (a) raw, Procrustes-aligned landmark coordinates (\emph{linear} shape descriptor, analogous to ordinary linear regression / PCA on $\R^{256}$ without the Riemannian log map) and (b) normalized CC volume/area, replicating the descriptor used by \citep{soysal2025machine}.

\subsection{Experiment 1: Age, Gender, and Diagnosis vs.\ Shape Evolution of the CC}

This experiment has several parts. We fit the regression using Algorithm~\ref{alg:georeg} to estimate the intercept $\bar{p}$ and slopes $v_k$. For each cohort and each nested model, we also fit a Euclidean benchmark (Linear / Multi-Linear Regression) on Procrustes-aligned landmarks as a comparison baseline. For every model, the reported $p$-value is obtained from a permutation test with $500$ permutations of the covariate rows.

\paragraph{CN-only cohort ($n=252$).}
The regression algorithm converged in $22$ iterations. Using the age covariate alone, we obtained a geodesic $R$-squared value of $0.1355$ with a permutation $p$-value of $0.0244$, confirming that age is a significant predictor of shape evolution in cognitively normal individuals. This result is comparable in magnitude to the value reported by Fletcher \citep{fletcher2013geodesic-regression} ( $R^2 \approx 0.12$ in the geodesic regression model). Adding sex as a second covariate produced a small improvement, with $R^2 = 0.1397$ and $p = 0.0244$, indicating that gender contributes little additional explanatory power beyond age in this cohort.

For the Euclidean benchmark on the same covariate sets, we obtained $R^2 = 0.1216$ for Age alone and $R^2 = 0.1291$ for Age+Sex. The geodesic model therefore outperformed the linear benchmark by $\Delta R^2 = +0.0139$ and $+0.0106$ respectively. This suggests that the shape evolution of the corpus callosum with age contains a non-linear component that the geodesic model captures but the flat Euclidean approximation does not.

\begin{figure}[H]
    \centering
    \includegraphics[width=0.48\linewidth]{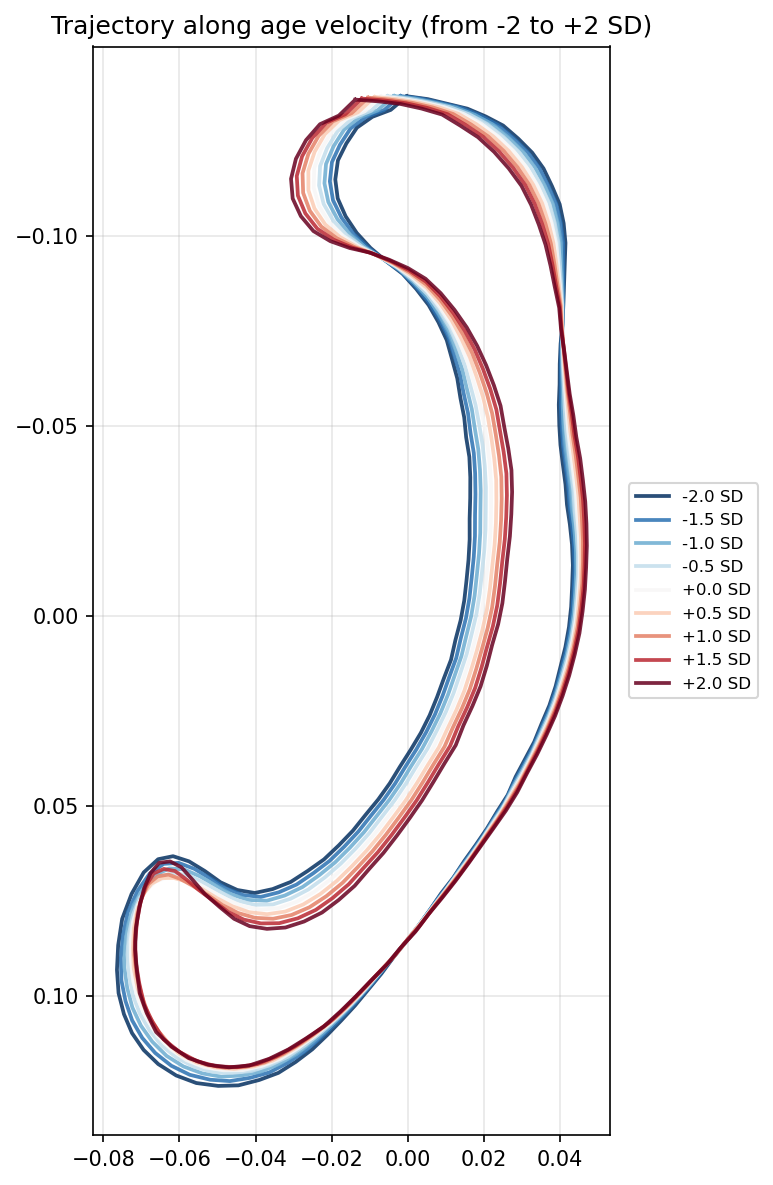}
    \caption{Trajectory along the age velocity from $-2$ to $+2$ SD.}
    \label{fig:shape_trajectory}
\end{figure}

\paragraph{AD-only cohort ($n=88$).}
The regression algorithm converged to very low $R$-squared values: $0.0040$ ($p=0.8293$) for Age alone and $0.0108$ ($p=0.9024$) for Age+Sex. Both are statistically insignificant. The Euclidean benchmark similarly produced $R^2 = 0.0025$ ($p=0.9154$) and $0.0089$ ($p=0.7512$). Taken together, these results indicate that chronological age and sex are not able to explain the shape evolution of the corpus callosum in Alzheimer's patients. Rather, the dominant driver of shape variation in this cohort appears to be the pathological progression of the disease itself, which is not captured by the demographic covariates alone.

\paragraph{Summary of the cohort comparison.}
The contrast between cohorts is clear: within the CN group alone, both
geodesic and linear regression capture a modest but highly significant
age-related shape signal; within the AD group alone, neither model finds
any significant signal. This dissociation is consistent with the
observation of Biegon et al.~\cite{biegon1994human}, who reported that
\emph{``selective changes within the corpus callosum accompany aging and
AD pathology''} and that \emph{``AD is accompanied by a large and
statistically significant reduction in the genu area''} relative to both
young and old controls. In other words, the morphological signature of
Alzheimer's disease is not a simple acceleration of the healthy aging
trajectory, but a partially distinct pathological process that
decorrelates callosal shape from chronological age. Across all cohorts
and all nested models, the geodesic regression either matches or
outperforms the Euclidean linear benchmark, with the largest geodesic
advantage appearing in the CN-only cohort where the shape evolution is
best characterized as a smooth non-linear trajectory in the curved shape
manifold.

\subsection{Experiment 2: Age Group Classification from Corpus Callosum}

The Cognitively Normal subgroup ($n = 252$) is divided into two age groups: \emph{young} (Age $\le 25$) and \emph{adults} (Age $> 25$). This split yields $n_{\text{young}} = 89$ and $n_{\text{adult}} = 163$ subjects. For each subject we compute its tangent-space deviation from the CN Fréchet mean via the Riemannian logarithm map $\Log_{[p]}([q])$, where $[p]$ is the Fréchet mean shape and $[q]$ is the subject's shape, both in Kendall shape space. We then concatenate the real and imaginary parts of the tangent vector, reduce with PCA, and train three classifiers — Logistic Regression, Random Forest, and SVM (RBF) — using stratified $5$-fold cross-validation. For each classifier, we report the mean accuracy and AUC across folds, together with permutation-based $p$-values computed from $60$ label permutations.

The results are summarized in Table~\ref{tab:classification}:
\begin{table}[h]
\centering
\begin{tabular}{lccc}
\hline
\textbf{Classifier} & \textbf{Accuracy} & \textbf{AUC} & \textbf{$p$-value} \\
\hline
Logistic Regression & $0.726 \pm 0.041$ & $\mathbf{0.791 \pm 0.049}$ & $0.016$ \\
Random Forest       & $0.722 \pm 0.044$ & $0.759 \pm 0.048$ & $0.016$ \\
SVM (RBF)           & $0.695 \pm 0.029$ & $0.772 \pm 0.044$ & $0.016$ \\
\hline
\end{tabular}
\caption{Young ($\le 25$) vs.\ adult ($> 25$) classification from corpus callosum shape in the CN cohort. All three classifiers achieve above-chance performance with statistically significant permutation $p$-values.}
\label{tab:classification}
\end{table}
Logistic Regression achieved the highest AUC ($0.791 \pm 0.049$), while SVM (RBF) achieved the highest accuracy ($0.726 \pm 0.041$). Random Forest performed comparably on accuracy ($0.722 \pm 0.044$) with slightly lower AUC ($0.759 \pm 0.048$). All three classifiers produced statistically significant results ($p = 0.016$ for both accuracy and AUC in every case), indicating that corpus callosum shape carries genuine and detectable age-group signal in the CN cohort.

\paragraph{Comparison with the volumetric approach of Soysal et al.}
The closest benchmark is \citep{soysal2025machine}, who applied machine learning to \emph{volumetric} measurements of the CC and its sub-regions (genu, body, splenium, left and right) from MRICloud segmentation. Their cohort ($n = 301$; ages $11$--$84$) was split into $52$ young (age $\le 25$) and $250$ adult subjects. Using $10$-fold cross-validation across $25$ algorithms, their best age-group classification reached $83.7\%$ accuracy with $\text{AUC} = 0.67$ (Fine Gaussian SVM).

Our Riemannian shape approach on the same task achieves $\text{AUC} = 0.791$ (Logistic Regression on the tangent-space embedding of the $128$-landmark midsagittal contour), a $+0.12$ improvement. The two experiments are not directly comparable, and three asymmetries should be kept in mind.

\textbf{Features.} Soysal et al.\ used six volumetric features derived from a full 3D segmentation of the structure ($\sim 10^5$ voxels). We use $128$ boundary points on a single 2D contour, reduced to $15$ PCA components. Our input is therefore strictly lower-dimensional: we sample a 2D closed curve, not a 3D volume. The AUC improvement reflects that the midsagittal contour is an information-rich region for this task, not that shape analysis dominates volumetric analysis in general.

\textbf{Classifier.} Their best was a Fine Gaussian SVM; ours was Logistic Regression. Kernel methods perform well in both settings, so the differing classifier choices are unlikely to explain the AUC gap.

\textbf{Modality.} Theirs is 3D volumetric; ours is 2D shape. A higher AUC from a 2D boundary-based method does not establish that shape analysis is intrinsically more powerful. It most likely reflects that (i) the dominant age-related shape variation is concentrated in the midsagittal contour, and (ii) the tangent-space representation aligns that variation with a linear classifier more directly than volumetric ratios do.
\subsection{Experiment 3: Age-Related Changes in Sub-Regions of the Corpus Callosum}

Anatomical deformation across the corpus callosum is spatially
heterogeneous and can obscure localized aging dynamics. We therefore
partition the $128$ landmarks into five anatomical sub-regions — Genu,
Anterior Body, Posterior Body, Isthmus, and Splenium — using arc-length
fractional boundaries along the midsagittal contour at $0.18 / 0.40 /
0.62 / 0.80$ from the genu tip, following the vertical reversed-C
convention in which the genu is the superior pole (higher $Y$) and the
splenium is the inferior pole (lower $Y$). Each subject's landmarks are
assigned by their own arc-length fraction relative to the detected genu
and splenium anchors, and a cohort-level majority vote across subjects
then fixes a single landmark-to-region map. The resulting assignment is
consistent across the cohort: Genu $23$, Anterior Body $29$, Posterior
Body $28$, Isthmus $23$, Splenium $25$ (sum $= 128$).

\paragraph{Regional velocity energy.}
We fit a single geodesic regression $\Exp_{[p_0]}(z_i \cdot v_\text{age})$ on the full 128-landmark shape across the CN cohort ($n = 252$). The fitted tangent $v_\text{age} \in T_{[p_0]}\text{CP}^{k-2}$ encodes shape motion per standard deviation of age, with $|v_\text{age}[j]|^2$ the local kinetic contribution of landmark $j$. Regional energy is
\begin{equation}
E_R = \sum_{j \in R} |v_\text{age}[j]|^2,
\end{equation}
with regional share $E_R / \sum_{R'} E_{R'}$. Because Kendall shape space has already removed translation, rotation, and size via Procrustes normalization, this energy reflects relative geometric deformation rather than raw size change.

\paragraph{Regional signed projection.}
To measure how strongly each region's shape tracks chronological age, we compute each subject's deviation from the Fréchet mean via the Riemannian logarithm map $w_i = \Log_{[p_0]}([q_i]) \in T_{[p_0]}\text{CP}^{k-2}$, and project it onto the fitted age velocity within region $R$:
\begin{equation}
s_i^{(R)} = \text{Re}\!\left( \sum_{j \in R} \overline{v_\text{age}[j]} \cdot w_i[j] \right).
\end{equation}
The scalar $s_i^{(R)}$ is positive when subject $i$'s region-$R$ shape has moved in the same direction as the fitted age velocity (``looks older''), and negative otherwise.

\paragraph{Results.}
Regional energy was heavily concentrated in the anterior corpus callosum: Genu $\mathbf{46.1\%}$, Anterior Body $23.3\%$, Splenium $12.6\%$, Posterior Body $12.2\%$, Isthmus $5.9\%$. The two anterior-most regions thus accounted for roughly $69\%$ of the total age-related shape change despite comprising only $\sim 41\%$ of the landmarks. Bootstrap resampling ($n = 200$) confirmed this ordering with narrow intervals: Genu $45.8\% \pm 2.4\%$, Anterior Body $23.2\% \pm 1.4\%$.

Correlating $s_i^{(R)}$ with chronological age, all five regions showed highly significant positive associations ($p < 10^{-12}$), but with markedly different strengths: Genu $r = 0.562$ ($R^2 = 0.316$), Anterior Body $r = 0.507$ ($R^2 = 0.257$), and the remaining three regions clustered at $r \approx 0.44$--$0.47$. The Genu's $R^2$ implies that roughly one third of the between-subject variation in how far the genu has moved along the age direction is explained by chronological age alone.

A related experiment that fits each region's shape against age in isolation yields a different ordering: the Anterior Body achieves the highest solo-fit $R^2$ ($0.101$), followed closely by the Genu ($0.100$) and the Isthmus ($0.085$). This is a shift in what is being measured. Energy and signed projection quantify how much age-driven deformation is concentrated in a region and how tightly that region's shape tracks age; the solo-fit metric instead rewards regions whose shape is \emph{predominantly} determined by age. The Anterior Body wins the solo fit because its geometry has fewer non-age modes of variation, not because it carries more absolute age signal.

Leave-one-region-out analysis confirmed the Genu's unique contribution: removing it reduced explained variance by the largest margin ($\Delta R^2 = +0.0304$), followed by the Splenium ($\Delta R^2 = +0.0175$); removing any other region did not degrade the model.

\begin{figure}[H]
    \centering
    \includegraphics[width=0.75\linewidth]{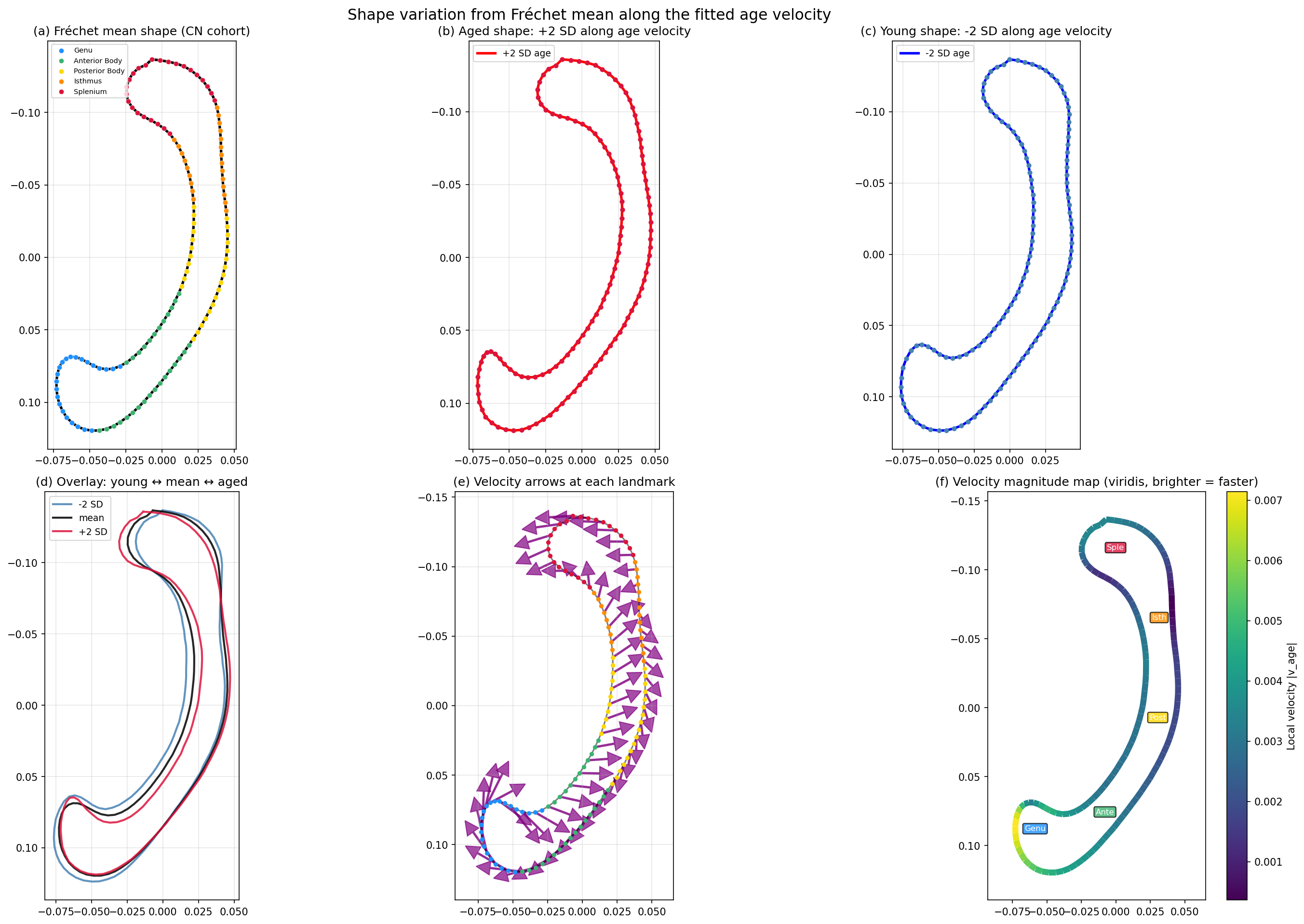}
    \caption{Regional Shape Variation with Age.}
    \label{fig:shape_trajectory}
\end{figure}

\paragraph{Relation to the literature.}
Our anterior-dominant ranking (Genu $>$ Anterior Body $>$ Splenium $>$ Posterior Body $>$ Isthmus) is consistent with a substantial literature documenting an anterior-to-posterior gradient of age-related vulnerability in the corpus callosum. Payas et al.~\cite{payas2026region} reported that \emph{``anterior and central pathways showed greater age-related change than posterior pathways''} and that \emph{``the forceps major demonstrated structural resilience with preserved surface area and microstructure.''} Liu et al.~\cite{liu2021connectivity}, analyzing 1,086 healthy adults across the lifespan, found that \emph{``anterior CC subregions exhibited larger microstructural and connectivity changes compared with the posterior CC subregions during aging''} and described an \emph{``overall `anterior-to-posterior' region-specific changing trend during aging.''} Fan et al.~\cite{fan2019age}, using high-gradient diffusion MRI, observed that age-related alterations in axon diameter index and density were \emph{``most pronounced in the genu''} and \emph{``relatively absent in the splenium.''} Sullivan et al.~\cite{sullivan2010quantitative} similarly stated that \emph{``a current consensus is that in adults advancing age affects anterior brain regions disproportionately more than posterior regions,''} with \emph{``FA lower and diffusivity higher in anterior than posterior fibers of regional paired comparisons (genu versus splenium).''} Biegon et al.~\cite{biegon1994human} reported that \emph{``AD is accompanied by a large and statistically significant reduction in the genu area,''} supporting the anterior specificity of callosal vulnerability to neurodegenerative processes. Our findings thus converge with this body of work: in the size-normalized shape representation, the genu and anterior body are the principal loci of age-related geometric deformation.

%%%%%%%%%%%%%%%%%%%%%%%%%%%%%%%%%%%%%%
\section{Conclusion and Future Work}
%==================================================================

We presented a Riemannian shape-space framework for analyzing age-related morphological change in the midsagittal corpus callosum. Working in Kendall shape space with $k = 128$ landmarks, we derived a multivariate geodesic regression with exact Riemannian gradients, benchmarked it against a corrected Euclidean baseline, and localized age-related deformation to anatomical sub-regions. Geodesic regression consistently matched or outperformed the linear benchmark, with the largest advantage in the cognitively normal cohort ($\Delta R^2 = +0.0139$). Regional energy was anterior-dominant — Genu $46.1\%$, Anterior Body $23.3\%$ — and signed projections confirmed the ordering (Genu $r = 0.562$; Anterior Body $r = 0.507$). This pattern aligns with the well-documented anterior-to-posterior gradient of age-related vulnerability in the corpus callosum~\cite{payas2026region, liu2021connectivity, fan2019age, sullivan2010quantitative}.

The most striking result is the cohort divergence: age explains $13.6\%$ of shape variance in the cognitively normal group but less than $0.5\%$ in Alzheimer's disease. The same covariate that drives a clear trajectory in healthy aging is essentially uninformative in AD.

\paragraph{Future work.}
This divergence suggests that deviations from the healthy aging trajectory may mark preclinical disease. Three steps would test this. First, project each subject onto the fitted CN trajectory and compute a per-subregion residual — five deviation scores per subject. Second, test whether these scores at baseline predict future conversion from CN to MCI or AD, and whether they correlate with amyloid burden. Third, compare deviation profiles between converters and non-converters to identify which subregions carry the earliest signal, using a cohort with longitudinal follow-up such as OASIS-3 or ADNI.

A complementary methodological direction is to extend the Kendall shape-space framework from the 2D midsagittal contour to the full 3D corpus callosum surface. A 3D representation would capture lateral extent and volumetric shape features that the midsagittal projection discards, at the cost of harder cross-subject correspondence and a less tractable geometry than $\CP^{k-2}$; geodesic regression on 3D surface shape spaces (or on stacked 2D contours across sagittal slices) would be the natural formulation. Whether this additional geometric information improves early-detection performance over the 2D contour remains an open empirical question.

%%%%%%%%%%%%%%%%%%%%%%%%%%%%%%%%%%%%%%

\bibliographystyle{plain}

\begin{thebibliography}{99}

\bibitem{dryden2016statistical}
I.~L. Dryden and K.~V. Mardia.
\newblock \emph{Statistical Shape Analysis, with Applications in R}.
\newblock 2nd edition, John Wiley \& Sons, Chichester, 2016.

\bibitem{docarmo1992riemannian}
M.~P. do Carmo.
\newblock \emph{Riemannian Geometry}.
\newblock Birkh\"auser, Boston, 1992.

\bibitem{fletcher2013geodesic-regression}
P.~T. Fletcher.
\newblock Geodesic regression and the theory of least squares on Riemannian manifolds.
\newblock \emph{International Journal of Computer Vision},
105(2):171--185, 2013.
\newblock doi:10.1007/s11263-012-0591-y.

\bibitem{hampel1998corpus}
H.~Hampel, S.~J. Teipel, G.~E. Alexander, B.~Horwitz,
D.~Teichberg, M.~B. Schapiro, and S.~I. Rapoport.
\newblock Corpus callosum atrophy is a possible indicator of region- and
cell type-specific neuronal degeneration in Alzheimer disease:
a magnetic resonance imaging analysis.
\newblock \emph{Archives of Neurology}, 55(2):193--198, 1998.
\newblock doi:10.1001/archneur.55.2.193.

\bibitem{karcher1977riemannian}
H.~Karcher.
\newblock Riemannian center of mass and mollifier smoothing.
\newblock \emph{Communications on Pure and Applied Mathematics},
30(5):509--541, 1977.
\newblock doi:10.1002/cpa.3160300502.

\bibitem{kendall1984shape}
D.~G. Kendall.
\newblock Shape manifolds, Procrustean metrics, and complex projective spaces.
\newblock \emph{Bulletin of the London Mathematical Society},
16(2):81--121, 1984.
\newblock doi:10.1112/blms/16.2.81.

\bibitem{lee2018introduction}
J.~M. Lee.
\newblock \emph{Introduction to Riemannian Manifolds}.
\newblock 2nd edition, Springer, Cham, 2018.

\bibitem{luders2010bell}
E.~Luders, K.~L. Narr, P.~M. Thompson, and A.~W. Toga.
\newblock Bell and halves: A review of size and shape differences
in the human corpus callosum.
\newblock \emph{Neuroscience Bulletin}, 26(6):464--472, 2010.

\bibitem{marcus2007oasis}
D.~S. Marcus, T.~H. Wang, J.~Parker, J.~G. Csernansky,
J.~C. Morris, and R.~L. Buckner.
\newblock Open Access Series of Imaging Studies (OASIS):
cross-sectional MRI data in young, middle aged, nondemented,
and demented older adults.
\newblock \emph{Journal of Cognitive Neuroscience},
19(9):1498--1507, 2007.
\newblock doi:10.1162/jocn.2007.19.9.1498.

\bibitem{soysal2025machine}
H.~Soysal, N.~Acer, M.~\"{O}zdemir, and K.~Gumus.
\newblock Machine learning for detecting age and sex effects on corpus callosum volume.
\newblock \emph{PeerJ Computer Science}, 11:e3179, 2025.

\bibitem{payas2026region}
A.~Payas, T.~Seber, F.~Cicek, H.~Kocaman, H.~Yildirim,
M.~Budak, and I.~U{\c{c}}ar.
\newblock Region-specific structural signatures of the corpus callosum
across adulthood: A tractography and machine learning study.
\newblock \emph{Brain Mechanisms}, 153:202543, 2026.
\newblock doi:10.1016/j.bramec.2026.202543.

\bibitem{liu2021connectivity}
Y.~Liu, C.-C.~H. Hsu, C.-C.~Huang, Y.~Zhang, J.~Zhao,
S.-J.~Tsai, L.-K.~Chen, C.-P.~Lin, and C.-Y.~Z. Lo.
\newblock Connectivity-based topographical changes of the corpus callosum
during aging.
\newblock \emph{Frontiers in Aging Neuroscience}, 13:753236, 2021.
\newblock doi:10.3389/fnagi.2021.753236.

\bibitem{fan2019age}
Q.~Fan, Q.~Tian, N.~A. Ohringer, A.~Nummenmaa, T.~Witzel,
S.~M. Tobyne, E.~C. Klawiter, C.~Mekkaoui, B.~R. Rosen,
L.~L. Wald, D.~H. Salat, and S.~Y. Huang.
\newblock Age-related alterations in axonal microstructure in the corpus
callosum measured by high-gradient diffusion MRI.
\newblock \emph{NeuroImage}, 191:325--336, 2019.
\newblock doi:10.1016/j.neuroimage.2019.02.036.

\bibitem{sullivan2010quantitative}
E.~V. Sullivan, T.~Rohlfing, and A.~Pfefferbaum.
\newblock Quantitative fiber tracking of lateral and interhemispheric
white matter systems in normal aging: Relations to timed performance.
\newblock \emph{Neurobiology of Aging}, 31(3):464--481, 2010.
\newblock doi:10.1016/j.neurobiolaging.2008.04.007.

\bibitem{biegon1994human}
A.~Biegon, J.~L. Eberling, B.~C. Richardson, M.~S. Roos,
S.~T. Wong, B.~R. Reed, and W.~J. Jagust.
\newblock Human corpus callosum in aging and Alzheimer's disease:
a magnetic resonance imaging study.
\newblock \emph{Neurobiology of Aging}, 15(4):393--397, 1994.
\newblock doi:10.1016/0197-4580(94)90070-1.


\end{thebibliography}

\end{document}